\documentclass[runningheads]{llncs}
\usepackage{graphicx}
\usepackage{amsmath,amssymb} 
\usepackage{color}
\usepackage{graphicx}
\usepackage{amsmath,amssymb} % define this before the line numbering.
\usepackage{color}
\usepackage{amsmath}
\usepackage{amssymb}
\usepackage{multirow}
\usepackage{float}
\usepackage{tabulary}
\usepackage{tabularx}
\usepackage[tight,normalsize,sf,SF]{subfigure}
\usepackage{color}
\usepackage{soul}
\usepackage{sidecap}
\usepackage{url}
\usepackage{enumitem}
\usepackage{hhline}
\usepackage{algorithm}
\usepackage{algpseudocode}
\usepackage{array}
\usepackage{flushend}
\usepackage{csquotes}
\newcolumntype{P}[1]{>{\centering\arraybackslash}p{#1}}
\DeclareMathSymbol{\Lambda}{\mathalpha}{operators}{3}
\DeclareMathSymbol{\Pi}{\mathalpha}{operators}{5}
\graphicspath{{./figs/}}
\DeclareGraphicsExtensions{.eps,.ps,.pdf,.png,.tiff}

\usepackage[table]{xcolor}
\usepackage{makecell,pict2e}
\usepackage{diagbox}

\begin{document}
\title{Contemplating Visual Emotions: Understanding and Overcoming Dataset Bias} 
% Replace with your title

\titlerunning{Contemplating Visual Emotions}
% Replace with a meaningful short version of your title
%
\author{Rameswar Panda\inst{1}\and
Jianming Zhang\inst{2} \and Haoxiang Li\inst{3} \and Joon-Young Lee\inst{2} \and Xin Lu\inst{2} \and 
Amit K. Roy-Chowdhury\inst{1}}

\authorrunning{R. Panda, J. Zhang, H. Li, J. Lee, X. Lu, and A. K. Roy-Chowdhury}
% Replace with shorter version of the author list. If there are more authors than fits a line, please use A. Author et al.
%

\institute{Department of ECE, UC Riverside. \email{\{rpand002@,amitrc@ece.\}@ucr.edu} \and Adobe Research. \email{\{jianmzha,jolee,xinl\}@adobe.com} \and Aibee. \email{hxli@aibee.com}}

\maketitle              % typeset the header of the contribution
\begin{abstract}
While machine learning approaches to visual emotion recognition offer great promise, current methods consider training and testing models on small scale datasets covering limited visual emotion concepts.
Our analysis identifies an important but long overlooked issue of existing visual emotion benchmarks in the form of dataset biases. We design a series of tests to show and measure how such dataset biases obstruct learning a generalizable emotion recognition model.
Based on our analysis, we propose a webly supervised approach by leveraging a large quantity of stock image data. 
Our approach uses a simple yet effective curriculum guided training strategy for learning discriminative emotion features. We discover that the models learned using our large scale stock image dataset exhibit significantly better generalization ability than the existing datasets without the manual collection of even a single label.
Moreover, visual representation learned using our approach holds a lot of promise across a variety of tasks on different image and video datasets.

\end{abstract}

\section{Introduction}

Recently, algorithms for object recognition and related tasks have become sufficiently proficient that new vision tasks beyond objects can now be pursued.
One such task is to \textit{recognize emotions expressed by images} which has gained momentum in last couple of years
in both academia and industries~\cite{you2016building,kim2017building,ng2015deep,peng2015mixed,you2015robust,chen2014deepsentibank}. 
Teaching machines to recognize diverse emotions is a very challenging problem with great application potential.

\begin{figure*} [t]
	\centering
	\begin{tabular}{c}
		\includegraphics[scale=0.35]{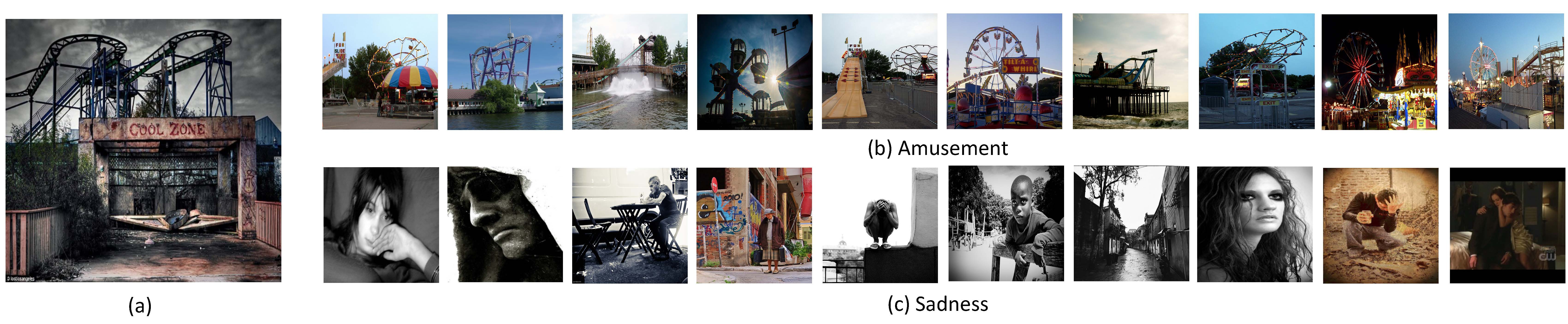}
	\end{tabular} 
	\caption
	{\scriptsize (a) An example image of an amusement park with negative emotion (sadness) (Source: Google Images). (b)-(c) Nearest neighbor images extracted from \enquote{amusement} and \enquote{sadness} category in the Deep Emotion dataset~\cite{you2016building}, which show a strong data bias. We use the pool5 features from our  ResNet-50 trained on Deep Emotion dataset to extract these nearest neighbor images.}
	\label{fig:DeepEmotionFig} 
\end{figure*}

Let us consider the image shown in Figure~\ref{fig:DeepEmotionFig}.a. Can you recognize the basic emotion expressed by this image? Practically, this should not be a difficult task as a quick glance can well reveal that the overall emotional impact of the image is negative (i.e., sadness) (9 out of 10 students in our lab made it correct!). In fact, this is the image of a Six Flags theme park at New Orleans which has been closed since Hurricane Katrina struck the state of Louisiana in August 2005.
\footnote{The image is taken from Google Images with the search keyword \textit{sad amusement park}. Source:
	\url{https://goo.gl/AUwoPZ}}

Intrigued, we decided to perform a toy experiment using Convolutional Neural Networks (CNNs) to recognize emotions. A ResNet-50~\cite{he2016deep} model that we trained on the current largest Deep Emotion dataset~\cite{you2016building} predicts an emotion of \enquote{amusement/joy} with 99.9\% confidence from the image in Figure~\ref{fig:DeepEmotionFig}.a. Why is this happening? Our initial investigation with the nearest neighbour images in Figure~\ref{fig:DeepEmotionFig}.b/c shows that the dataset bias appears to be the main culprit. Specifically, the Deep Emotion dataset~\cite{you2016building} suffers from two types of biases. The first is the positive set bias, which makes the \emph{amusement} category in the dataset full of photos of amusement parks (see Figure~\ref{fig:DeepEmotionFig}.b). This is due to the lack of diversity in visual concepts when collecting the source images. The second is the negative set bias, where the rest of the dataset does not well represent the rest of the world, i.e., no images of sad park in the dataset (see Figure~\ref{fig:DeepEmotionFig}.c).   

In this paper, instead of focusing on beating the latest benchmark numbers on the latest dataset, we take a step back and pose an important question: \textit{how well do the existing datasets stack up overall in the emerging field of visual emotion recognition?} We first conduct a series of tests including a novel correlation analysis between emotion and object/scene categories to analyze the presence of bias in existing benchmarks.
We then present a number of possible remedies, mainly proposing a new weakly-labeled large-scale emotion dataset collected from a stock website and a simple yet effective curriculum guided training strategy for learning discriminative features. 
Our systematic analysis, which is first in emotion recognition, will provide insights to the researchers working in this area to focus on the right training/testing protocols and more broadly simulate discussions in the community regarding this very important but largely neglected issue of dataset bias in emotion recognition. 
We also hope our efforts in releasing several emotion benchmarks in this work will open up avenues for facilitating progress in this emerging area of computer vision.~\footnote{All our datasets, models and supplementary material are publicly available on our project page: \url{https://rpand002.github.io/emotion.html}}

The key takeaways from this paper can be summarized as follows:

\begin{itemize} 
	\item \textbf{Existing visual emotion datasets appear to have significant bias.} We conduct extensive studies and experiments for analyzing emotion recognition datasets (Sec.~\ref{sec:Bias}).
	Our analysis reveals the presence of significant biases in current benchmark datasets and calls for rethinking the current methodology for training and testing emotion recognition models.
	\item \textbf{Learning with large amounts of web data helps to alleviate (at least minimize) the effect of dataset bias. } We show that models
	learned using large-scale stock data exhibit
	significantly better generalization ability while testing on new unseen datasets (Sec.~\ref{sec:revisit}). 
	We further propose a simple yet effective curriculum guided training strategy (Sec.~\ref{sec:Curriculum}) for learning discriminative emotion features that achieves state-of-the-art performance on various tasks across different image and video datasets (Sec.~\ref{sec:features}). For example, we show improved performance ($\sim$3\% in top-5 mAP) of a state-of-the-art video summarization algorithm~\cite{panda2017collaborative} by just plugging in our emotion features. 
	\item \textbf{New Datasets.} We introduce multiple image emotion datasets collected from different sources for model training and testing. Our stock image dataset is one of the largest in the area of visual emotion analysis containing about 268,000 high quality stock photos across 25 fine-grained emotion categories. 
\end{itemize}

\section{Related Work}
\label{sec:Related Work}

\textbf{Emotion Wheels.} Various types of emotion wheels have been studied in psychology, e.g., Ekman’s emotions~\cite{1992argument} and Plutchik's emotions~\cite{plutchik1980emotion}. Our work is based on the popular Parrott's wheel of emotions~\cite{parrott2001emotions} which organizes emotions in the form of a tree with primary, secondary and tertiary emotions.
This hierarchical grouping is more interpretable and can potentially help to learn a better recognition model by leveraging the structure.

\noindent\textbf{Image Emotion Recognition.} A number of prior works studying visual emotion recognition focus on analyzing facial expressions~\cite{emotic_cvpr2017,du2014compound,fabian2016emotionet,eleftheriadis2015discriminative,eleftheriadis2016joint,soleymani2016analysis,du2014compound,chu2017selective}. Specifically, these works mainly predict emotions for images that involve a clear background with people as the primary subject. Predicting emotions from user-generated videos~\cite{kahou2013combining,jou2014predicting,xu2016video}, social media images~\cite{wu2017inferring,xu2016video,wang2015modeling} and artistic photos~\cite{zhao2014exploring,alameda2016recognizing} are also some recent trends in emotion recognition.
While these approaches have obtained reasonable performance on such controlled emotion datasets, they have not yet considered predicting emotions from natural images as discussed in this paper.
%(See Fig.~\ref{fig:BiasFig} for example)
Most related to our work along the direction of recognizing emotions from natural images are the works of~\cite{you2016building,machajdik2010affective,kim2017building,peng2015mixed} which predict emotions from images crawled from Flickr and Instagram. As an example, the authors in~\cite{you2016building} learn a CNN model to recognize emotions in natural images and performs reasonably well on the Deep Emotion dataset~\cite{you2016building}. However, it requires expensive human annotation and is difficult to scale up to cover the diverse emotion concepts.
Instead, we focus on webly supervised learning of CNNs which can potentially avoid (at least minimize) the dataset design biases by utilizing vast amount of weakly labeled data from diverse concepts.

\noindent\textbf{Webly Supervised Learning.} There is a continued interest in the vision community on learning recognition models directly from web data since images on the web can cover a wide variety of visual concepts and, more importantly, can be used to learn computational models without using instance-level human annotations~\cite{li2017webvision,chen2015webly,divvala2014learning,joulin2016learning,gan2016webly,liang2016learning,gan2016you,sukhbaatar2014learning,krause2016unreasonable,liang2016exploiting}. While the existing works have shown advantages of using web data by either manually cleaning the data or developing a specific mechanism for reducing the noise level, we demonstrate that noisy web data can be surprisingly effective with a curriculum guided learning strategy for recognizing fine-grained emotions from natural images.

\noindent\textbf{Curriculum Learning.} Our work is related to curriculum learning~\cite{lee2011learning,dong2017multi,zhang2017curriculum,gao2017demand,pentina2015curriculum,bengio2009curriculum} that learns a model by gradually including easy to complex samples in training so as to increase the entropy of training samples. However, unlike these prior works that typically focus on the evolution of the input training data, our approach focuses on the evolution of the output domain, i.e., evolution of emotion categories from being easy to difficult in prediction.

\noindent\textbf{Hierarchical Recognition.} Category hierarchies have been successfully leveraged in several recognition tasks: image classification~\cite{yan2015hd,xiao2014error,griffin2008learning,li2010building,cesa2006incremental,deng2012hedging}, object detection~\cite{deng2011fast,marszalek2007semantic}, image annotation~\cite{tousch2012semantic}, and concept learning~\cite{jia2013visual} (see~\cite{silla2011survey} for an overview). 
CNN based methods~\cite{srivastava2013discriminative,yan2015hd,xiao2014error,wang2015multiple} have also used class hierarchy for large scale image classification. Unlike these methods that mostly use clean manually labeled datasets to learn the hierarchy, we adopt an emotion hierarchy from psychology~\cite{parrott2001emotions} to guide the learning with noisy web data. Our basic idea is that the emotion hierarchy can provide guidance for learning more difficult tasks in a sequential manner and also provide regularization for label noises.

\section{Understanding Bias in Emotion Datasets}
\label{sec:Bias}

\textbf{Goal.} Our main goal in this section is to identify, show and measure dataset bias in existing emotion recogntion datasets using a series of tests.

\noindent\textbf{Datasets.} We pick three representative datasets including one newly created by us: (1) Deep Sentiment~\cite{you2015robust} dataset containing 1269 images from Twitter, (2) the current largest Deep Emotion dataset~\cite{you2016building}, (3) our Emotion-6 dataset of 8350 images (\emph{anger}: 1604, \emph{fear}: 1280, \emph{joy}: 1964, \emph{love}: 724, \emph{sadness}: 2221, \emph{surprise}: 557) labeled by five human subjects from intially 150K images collected from Google and Flickr (see supp). 
Our main motivation on creating Emotion-6 dataset is to repeat the standard data collection/annotation protocol used by existing works~\cite{you2016building,you2015robust} and see how well it performs regarding the dataset biases.

\begin{figure*} [t]
	\begin{center}
		\includegraphics[scale=0.32]{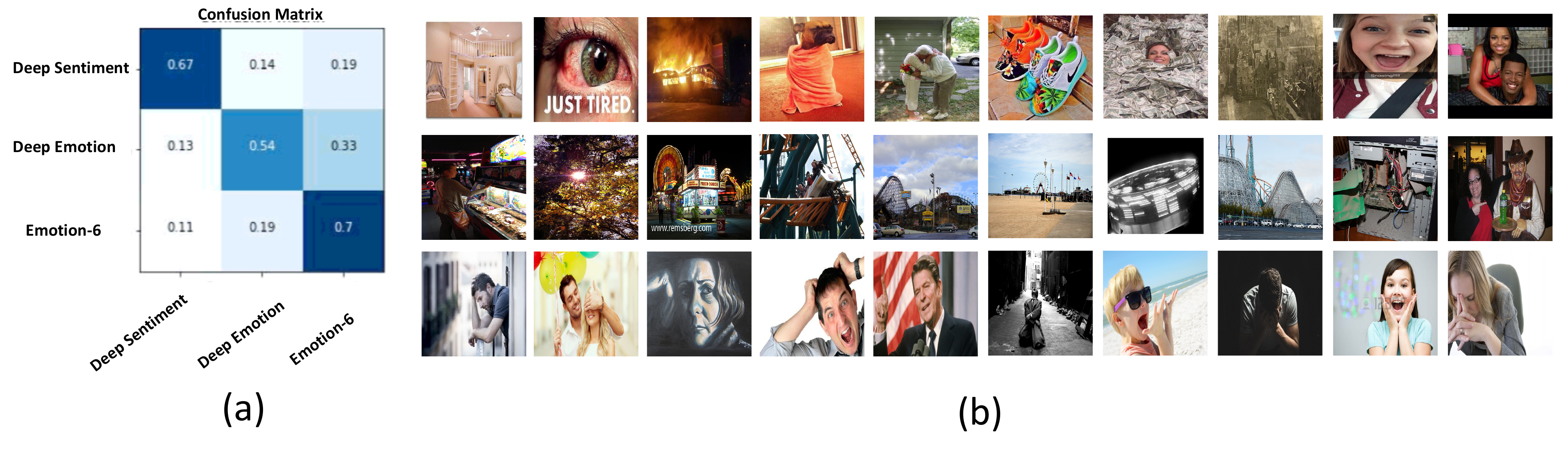} 
	\end{center} 
	\caption{\scriptsize (a) Confusion matrix, (b) From top to bottom, depicted are examples of high confident correct predictions from Deep Sentiment, Deep Emotion and Emotion-6 datasets respectively. }
	\label{fig:NameDataset}
\end{figure*}

\noindent\textbf{Test 1. Name That Dataset Game.} 
With the aim of getting an initial idea on the relation among different datasets, we start our analysis by running \textit{Name That Dataset Game} as in~\cite{torralba2011unbiased}. We randomly sample 500 images from the training portions of each of the three datasets and train a 3-class linear classifier over the ResNet-50 features. We then test on 100 random images from each of the test sets and observe that the classifier is reasonably good at telling different datasets apart, giving 63.67\% performance. The distinct diagonal in confusion matrix (Figure~\ref{fig:NameDataset}.a) shows that these datasets possesses an unique signature leading to the presence of bias. For example, visually examining the high confidence correct predictions from the test set in Figure~\ref{fig:NameDataset}.b indicates that Deep Emotion dataset has a strong preference for outdoor scenes mostly focusing on parks (2nd row), while Emotion-6 tend to be biased toward images where a single object is centered with a clean background and a canonical viewpoint (3rd row).

\noindent\textbf{Test 2. Binary Cross-Dataset Generalization.}
Given all three datasets, we train a ResNet-50 classifier to show cross-dataset generalization i.e., training on one dataset while testing on the other. 
For both Deep Emotion and Emotion-6, we randomly sample 80\% of images for training and keep rest 20\% for testing, while on Deep Sentiment, we use 90\% of images for the training and keep the rest for testing, as in~\cite{you2015robust}.
Since, exact emotion categories can vary from one dataset to another, we report binary classification accuracies (positive vs negative) which are computed by tranforming the predicted labels to two basic emotion categories, following Parrott's emotional grouping~\cite{parrott2001emotions}. We call this \emph{Binary Cross-Dataset Generalization Test}, as it asks the CNN model to predict the most trivial basic emotion category from an image. If a model cannot generalize well in this simple test, it will not work on more fine-grained emotion categories. Moreover, the binary generalization test only involves minimum post-processing of the model predictions, so it can evaluate different datasets more fairly.

\begin{table} [t]
	\centering
	\scriptsize
	\caption{\scriptsize Binary Cross-Dataset Generalization. Diagonal numbers refer to training and testing on same dataset while non-diagonal numbers refer to training on one dataset and testing on others. \% Drop refers to the performance drop across the diagonal and the average of non-digonal numbers.} 
	\begin{tabular}{|l||*{4}{c|}}\hline
		\label{tab:cross}
		\centering
		\tiny
		\backslashbox{Train on:}{Test on:}
		&\makebox[6.2em]{\tiny Deep Sentiment} &\makebox[6em]{\tiny Deep Emotion}&\makebox[6em]{\tiny Emotion-6}&\makebox[4em]{\tiny \% Drop}
		\\\hline\hline
		\tiny Deep Sentiment & \scriptsize \textcolor{red}{78.74} & \scriptsize 68.38 &\scriptsize 49.76 &\scriptsize \textbf{24.98} \\\hline
		\tiny Deep Emotion& \scriptsize 61.41 & \scriptsize \textcolor{red}{84.81} &\scriptsize 69.22 & \scriptsize \textbf{22.99}\\\hline
		\tiny Emotion-6 &\scriptsize 54.33 &\scriptsize 64.28 &\scriptsize \textcolor{red}{77.72} &\scriptsize \textbf{23.69} \\\hline
	\end{tabular} 
\end{table}

Table~\ref{tab:cross} shows a summary of results. From Table~\ref{tab:cross}, the following observations can be made: (1) As expected, training and testing on the same dataset provides the best performance on all cases (marked in red). (2) Training on one dataset and testing on the other shows a significant drop in accuracy, for instance, the classifier trained on Deep Emotion dataset shows a average drop of 22.99\% in accuracy while testing on other two datasets. Why is this happening?
Our observations suggest that the answer lies in the emotion dataset itself: it's size is relatively small, which results in the positive set bias due to the lack of diversity in visual concepts. 
As a result, models learned using such data essentially memorize all it's idiosyncrasies and lose the ability to generalize.

\noindent\textbf{Test 3. Quantifying Negative Bias.}
We choose three common emotion categories across Deep Emotion and Emotion-6 datasets (\textit{anger}, \textit{fear} and \textit{sadness}) to measure negative set bias in different datasets. For each dataset, we train a binary classifier (e.g., anger vs non-anger) on its own set of positive and negative instances while for testing, the positives come from that dataset, but the negatives come from other datasets. We train the classifiers on 500 positive and 2000 negative images randomly selected from each dataset. Then for testing, we use 200 positive and 4000 negative images from other datasets. 

Table~\ref{tab:neg} summarizes the results. For both datasets, we observe a significant decrease in performance (maximum of about 25\% for Deep Emotion dataset on \textit{sadness} emotion), suggesting that some of the new negative samples coming from other datasets are confused with positive examples. This indicates that rest of the dataset does not well represent the rest of the visual world leading to overconfident, and not very discriminative, classifiers.

\begin{table} [t]
	\centering
	\scriptsize
	\caption{\scriptsize Quantifying Negative Bias. Self refers to testing on the original test set while Others refer to the testing on a set where positives come from the original dataset but negatives come from the other. \% Drop refers to the performance drop across the self and others. Values in Others represent the average numbers. WEBEmo refers to our released dataset that we will discuss in next section.} 
	\begin{tabular}{|l||l||c|c||c|}\hline
		\label{tab:neg}
		\centering
		\tiny
		Task	&\backslashbox{-ve set:}{+ve set:}
		&\makebox[6.2em]{\tiny Deep Emotion} &\makebox[6em]{\tiny Emotion-6}&\makebox[6em]{\tiny WEBEmo}
		\\\hline\hline
		\tiny anger vs non-anger	& \tiny Self/Others/\% Drop & \scriptsize 90.64/78.98/\textbf{12.86} & \scriptsize 92.40/83.56/\textbf{9.57} &\scriptsize 83.90/83.37/\textbf{0.63}  \\\hline
		\tiny fear vs non-fear	& \tiny Self/Others/\% Drop & \scriptsize 85.95/80.77/\textbf{6.05} & \scriptsize 81.14/76.02/\textbf{2.56} &\scriptsize 82.97/84.79/\textbf{-2.19} \\\hline
		\tiny sadness vs non-sadness	& \tiny Self/Others/\% Drop &\scriptsize 81.90/61.35/\textbf{25.09} &\scriptsize 89.20/82.07/\textbf{7.99} &\scriptsize 89.89/90.55/\textbf{-0.73} \\\hline
	\end{tabular} 
\end{table}

\noindent\textbf{Test 4. Correlation Analysis with Object/Scene Categories.}
Given existing object/scene recognition models, the objective of this test is to see how well emotions are correlated with object/scene categories and whether analyzing the correlations can help to identify the presence of bias in emotion datasets. We use ResNet-50 pre-trained on ImageNet and ResNet-152 pre-trained on Places365 as object and scene recognition models respectively. We start our analysis by predicting object/scene categories from images of three common emotion categories used in previous task. We then select top 200 most occuring object/scene categories from each emotion class and compute the conditional entropy of each object/scene category across positive and negative set of a specific emotion.
Mathematically, given an object/scene category $c$ and emotion category $e$, we compute the conditional entropy as $\mathcal{H}(Y|X=c)= -\sum_{y\epsilon\{e_p,e_n\}}p(y|X=c)\text{log}p(y|X=c)$, where $e_p$ and $e_n$ represent the positive and negative set of emotion $e$ respectively (e.g., anger and non-anger).
More number of object/scene categories with zero conditional entropy will most likey lead to a biased dataset as it shows the presence of these object/scene categories in either positive or negative set of an emotion resulting in an unbalanced representation of the visual world (Figure~\ref{fig:DeepEmotionFig}). 

\begin{figure*} [t]
	\begin{center}
		\begin{tabular}{cc}
			\includegraphics[scale=0.086]{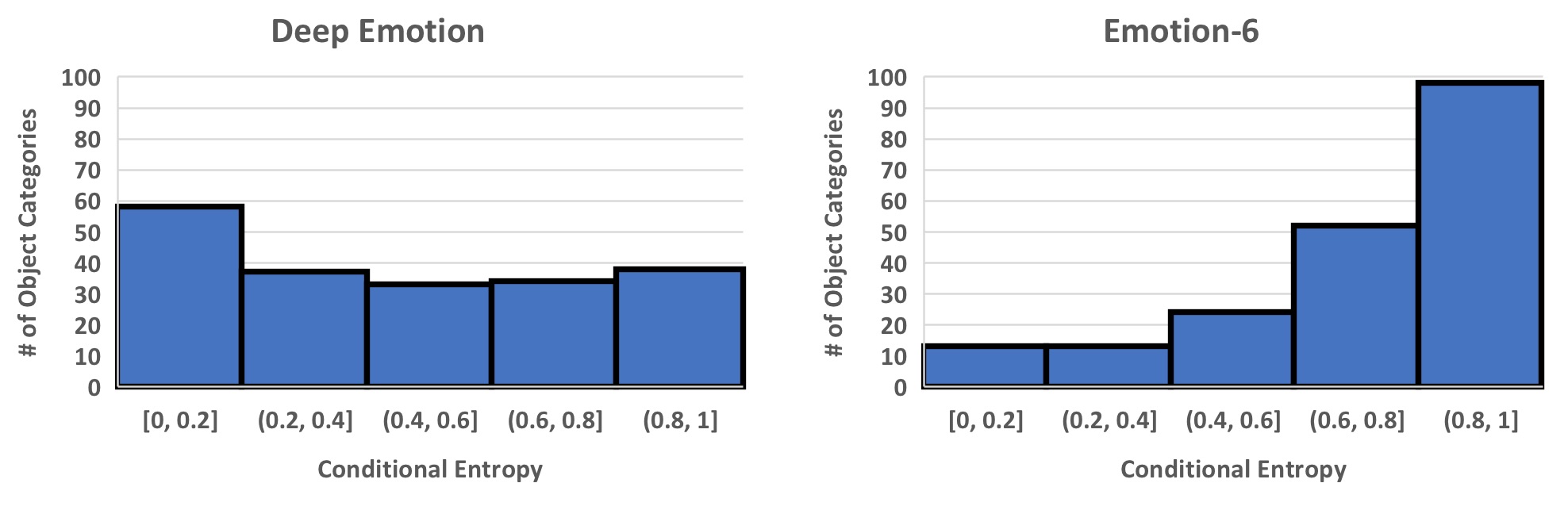} &
			\includegraphics[scale=0.086]{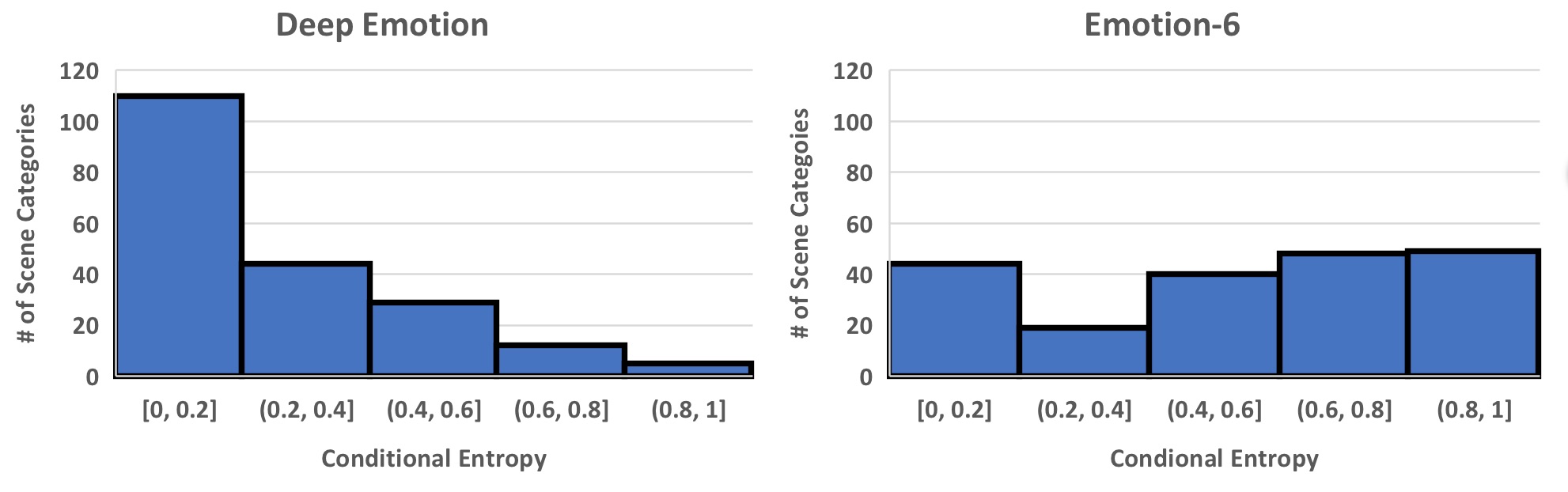} \\
			{\scriptsize (a) Object Categories for \textit{Sadness} Emotion} & \scriptsize{ (b) Scene Categories for \textit{Anger} emotion.}  \\
		\end{tabular} \vspace{-7mm}
	\end{center} 
	\caption{\scriptsize Distribution of object/scene categories w.r.t conditional entropy. (a) objects in \textit{sadness} emotion, (b) scenes in \textit{anger} emotion. Both datasets show a strong presence of bias.} 
	\label{fig:Obj_Scene} 
\end{figure*}

Figure~\ref{fig:Obj_Scene} shows the distribution of object/scene categories w.r.t conditional entropy for both Deep Emotion and Emotion-6 datasets. While analyzing correlations between objects and \textit{sadness} emotion in Figure~\ref{fig:Obj_Scene}.a, we observe that about 30\% of object categories (zero conditional entropy) are only present in either sadness or non-sadness category and then further examining these categories, we find most of them will lead to a dataset bias (see supp). 
For example, objects like balloon, candy store and parachute are only present in negative set of \textit{sadness}.
Categories like balloon are strongly related to happiness, but still there should be a few negative balloon images such as sad balloon in the negative set\footnote{For example, see: ~\url{https://tinyurl.com/yazvkjmv}}. Completely missing the negative balloon images will lead to dataset bias. 
Emotion-6 appears to be less biased compared to Deep Emotion but still it has 25\% of object categories in the entropy range of [0,0.5]. 
Similarly, on analyzing scene categories for \textit{anger} emotion in Fig.~\ref{fig:Obj_Scene}.b, we see that both datasets are biased towards to specific scene categories, e.g., for Deep Emotion, about 55\% of scene categories have zero conditional entropy while about 20\% of categories have zero entropy in Emotion-6. More results are included in the supplementary.

Our main conclusions from these series of tests indicate that despite all three datasets being collected from Internet and labeled using a similar paradigm involving multiple humans, they appear to have strong bias which severly obstruct learning a generalizable recognition model.

\section{Curriculum Guided Webly Supervised Learning}
\label{sec:Curriculum}

\textbf{Goal.} The main goal of this section is to present possible remedies to the dataset bias issues described above, mainly proposing a large-scale web emotion database, called \textbf{WEBEmo} and an effective curriculum guided strategy for learning discrimative emotion features. Our basic idea is that we can potentially avoid (at least minimize) the effect of dataset design biases by exploiting vast amount of freely available web data covering a wide variety of emotion concepts.

\noindent\textbf{Emotion Categories.} Emotions can be grouped into different categories. Most prior works only consider a few independent emotion categories, e.g., Ekmas's six emotions~\cite{1992argument} or Plutchik's eight emotion categories~\cite{plutchik1980emotion}. Instead, we opt for Parrott's hierarchical model of emotions~\cite{parrott2001emotions} for two main advantages. First, by leveraging this hierarchy with associated lists of keywords, we are able to allieviate the search engine bias by diversifying the image search. Second, we are able to learn discriminative features by progressively solving different tasks.

\begin{SCfigure} 
	\label{fig:DistWEBEmoFig} 
	\caption{\scriptsize Category-wise distribution of images in \textbf{WEBEmo} dataset. The are more than 30K images on \textit{cheerfulness} category while only 629 images are there on \textit{enthrallment} emotion category.  Categories are sorted according to the number of images in corresponding category, from the highest (left) to the lowest (right). Best viewed in enlarged version.}
	\includegraphics[scale=0.22]{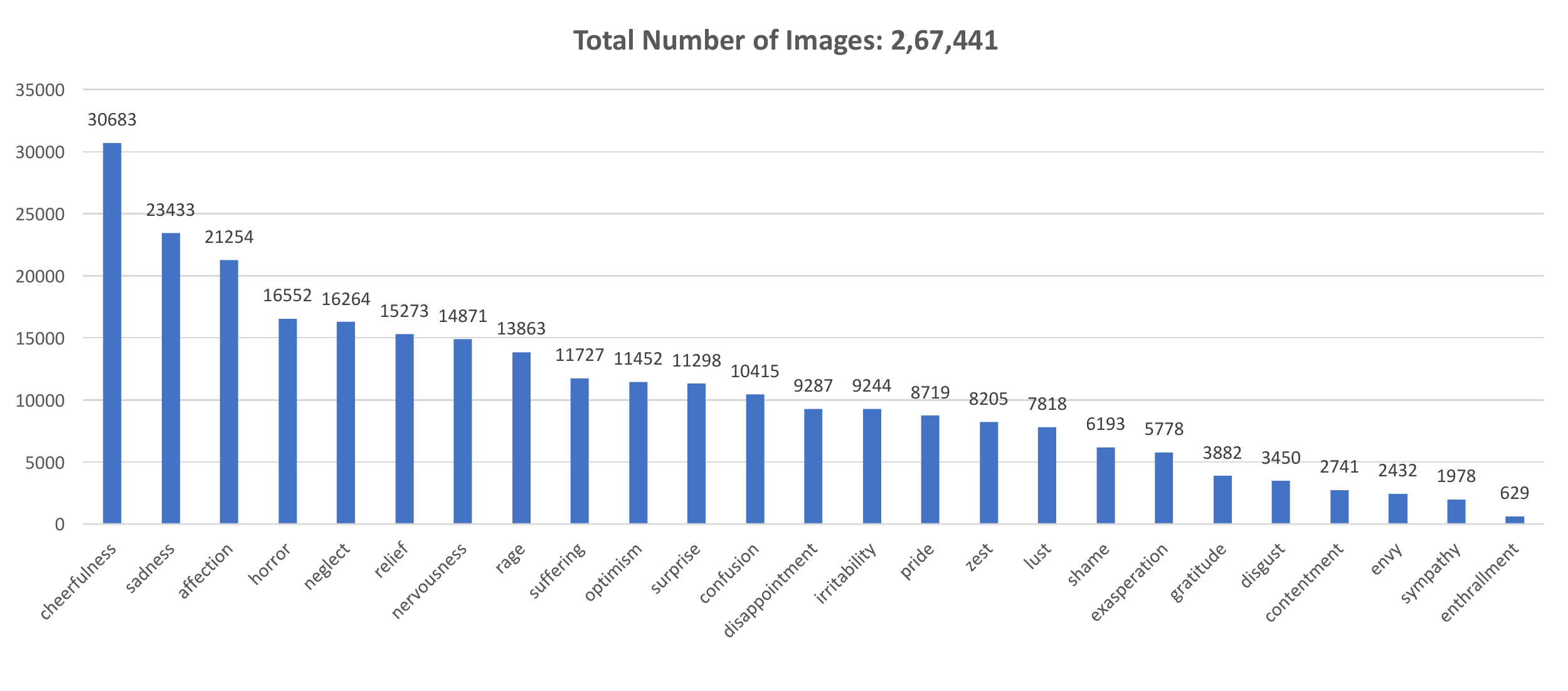}
\end{SCfigure}

Following~\cite{parrott2001emotions}, we design a three-level emotion hierarchy, starting from two basic categories (\emph{positive} and \emph{negative}) at level-1, six categories (\emph{anger}, \emph{fear}, \emph{joy}, \emph{love}, \emph{sadness}, and \emph{surprise}) at level-2 to 25 fine-grained emotion categories at level-3 (see Figure~\ref{fig:DistWEBEmoFig} for all categories). Note that while data-driven learning~\cite{verma2012learning,li2010building} can be used for constructing such hierarchy, we chose to design it following prior psychological studies~\cite{parrott2001emotions} as emotion has been well studied in psychology.

\noindent\textbf{Retrieving Images from the Web.}
We use a stock website to retrieve web images and use those images without any additional manual labeling. Below, we provide a brief description of the dataset and refer to supplementary for details.

\begin{figure} [t]
	\begin{tabular}{c}
		\includegraphics[scale=0.15]{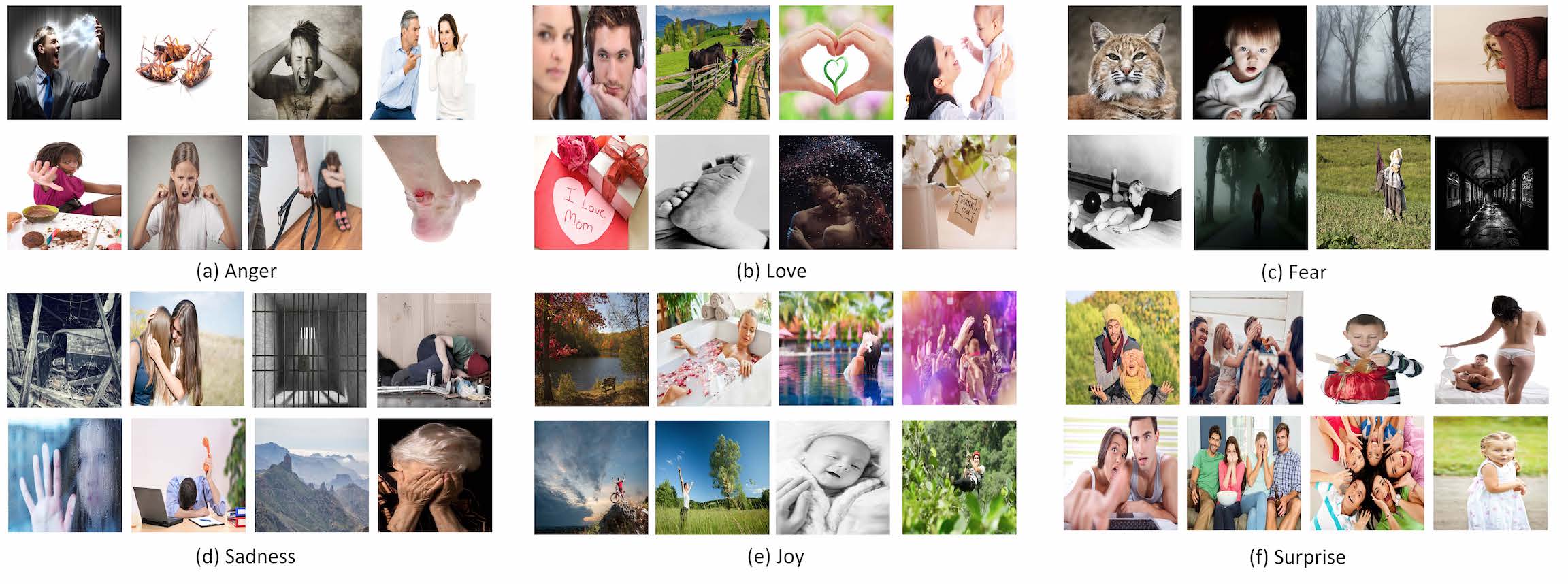}
	\end{tabular} 
	\caption
	{\scriptsize Sample images from our \textbf{WEBEmo} dataset across six secondary emotion categories. These images cover a wide range of visual concepts. Best viewed in color.} 
	\label{fig:AdobeStockFig} 
\end{figure}

To collect web images for emotion recognition, we follow~\cite{parrott2001emotions} to assemble a list of keywords (shown in supp) for each 25 fine-grained emotions, focusing on diverse visual concepts (see Figure~\ref{fig:AdobeStockFig}). We then use the entire list of keywords to query a stock site and retrieve all the images ($\sim$10,000) together with their tags returned for each query. In this way, we are able to collect about 300,000 weakly labeled images, i.e., labeled by the queries. We then remove images with non-English tags and also use captions with top-5 tags to remove duplicate images. After deduplication, we ended up with about 268,000 high-quality stock images.
Figure~\ref{fig:DistWEBEmoFig} shows category-wise distribution of images in \textbf{WEBEmo} dataset.  
The total number of images in our \textbf{WEBEmo} dataset is about 12 times larger than the current largest Deep Emotion dataset~\cite{you2016building}.

\noindent\textbf{Curriculum Guided Training.} Our goal is to learn discriminative features for emotion recognition directly using our \textbf{WEBEmo} database. While it seems that one can directly train a CNN with such data, as in~\cite{krause2016unreasonable} for image classificaton, we found it is extremely hard to learn good features for our task, as emotions are intrinsically fine-grained, ambigious, and web data is more prone to label noise.
However, as shown in psychology~\cite{parrott2001emotions}, emotions are organized in a hierarchy starting from basic emotions like postive or negative to more fine-grained emotions like affection, contentment, optimism and exasperation, etc. Categorizing images to two basic emotions is an easier task compared to categorizing images to such fine-grained emotions. So, what we want is an approach that can learn visual representation in a sequential manner like we humans normally learn difficult tasks in an organized manner.

Inspired by curriculum learning~\cite{bengio2009curriculum} and the emotion wheel from psychology~\cite{parrott2001emotions}, we develop a curriculum guided strategy for learning discriminative features in a sequential manner. Our basic idea is to gradually inject the information to the learner (CNN) so that in the early stages of training, the coarse-scale properties of the data are captured while the finer-scale characteristics are learned in later stages. Moreover, since the amount of label noise is likey to be much less in coarse categories, it can produce regularization effect and enhance the generalization of the learned representations.

Let $C$ be the set of fine-grained emotion categories (= 25 in our case) and $k \in \{1\dots K\}$ be the different stages of training. 
Assume $C_K=C$ is the fine-grained emotion categories that we want to predict; that is, our target is to arrive at the prediction of these emotion labels at the final stage of learning $K$.
In our curriculum guided learning, we require a stage-to-stage emotional mapping operator $\mathcal{F}$ which projects $C_k$, the output labels at stage $k$, to a lower-dimensional $C_{k-1}$ which is easier to predict compared to the prediction of $C_k$ labels.
We follow the Parrott's emotion grouping~\cite{parrott2001emotions} as the mapping operator that groups $C_K$ categories into six secondary and two primary level emotions as described earlier. Specifically, a CNN (pre-trained on ImageNet) is first fine-tuned with 2 basic emotions (positive/negative) at level-1 and then it serves to initialize a second one that discriminates
six emotion categories at level-2 and the process is finally repeated for 25 fine-grained emotion categories at level-3. 

\section{Experiments}
\label{sec:Experiments}

\textbf{Goal.} We perform rigorous experiments with the following two main objectives: 

(a) How well our newly introduced \textbf{WEBEmo} dataset along with the curriculum guided learning help in reducing the dataset bias? (Sec.~\ref{sec:revisit})

(b) How effective our visual representation learned using \textbf{WEBEmo} dataset in recognizing both image and video emotions? Do emotion features benefit other visual analysis tasks, say video summarization? (Sec.~\ref{sec:features})   

\noindent\textbf{Implementation Details.} All the networks are trained using the Caffe toolbox~\cite{jia2014caffe}. We choose ResNet-50~\cite{he2016deep} as our default deep network and initialize from an ImageNet checkpoint while learning using web data~\cite{sun2017revisiting}. During training, all input images are resized to 256 $\times$ 256 pixels and then randomly cropped to 224 $\times$ 224. We use batch normalization after all the convolutional layers and train using stochastic gradient descent with a minibatch size of 24, learning rate of 0.01, momentum of 0.9 and weight decay of 0.0001. 
We reduce the learning rate to its $\frac{1}{10}$ while making transition in our curriculum guided training.

\subsection{Revisiting Dataset Bias with Our Approach}
\label{sec:revisit}

\noindent\textbf{Experiment 1: Quantifying Negative Bias.}
We use the same number of images (total 2500 for training and 4200 for testing) and follow the exact same testing protocol mentioned in Sec.~\ref{sec:Bias}: Test 2 to analyze negative bias on our \textbf{WEBEmo} dataset.  
Table~\ref{tab:neg} shows that classifiers trained on our dataset do not seem to be affected by a new external negative set across all three emotion categories (see right most column in Table~\ref{tab:neg}). This is because \textbf{WEBEmo} dataset benefits from a large variability of negative examples and hence more comprehensively represent
the visual world of emotions.

\begin{SCfigure}
	\label{fig:Obj_Scene1} 
	\caption{\scriptsize Distribution of object/scene categories w.r.t conditional entropy on \textbf{WEBEmo} dataset. (a) objects in \textit{sadness}, (b) scenes in \textit{anger} emotion.}
	\includegraphics[scale=0.1]{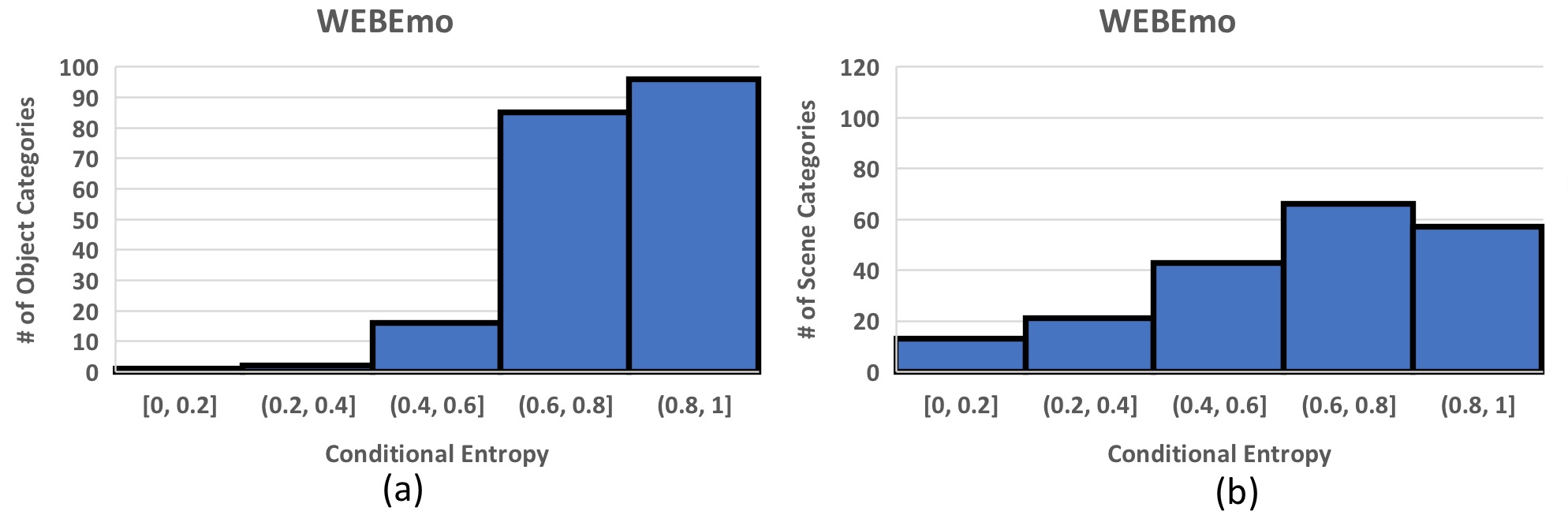} 
\end{SCfigure} 

\begin{table*} [t]
	\scriptsize
	\centering
	\caption{\scriptsize Cross-Dataset Generalization. \enquote{Self} refers to training and testing on same dataset and \enquote{Mean Others} refers to the mean performance on all others. Model trained using curriculum guided webly supervised learning generalizes well to other datasets.} 
	\begin{tabular}{|l||*{6}{c|}}\hline
		\label{tab:cross1}
		\centering
		\tiny
		\backslashbox{Train on:}{Test on:}
		&\makebox[6em]{\tiny Deep Sentiment} &\makebox[5.2em]{\tiny Deep Emotion} &\makebox[4.2em]{\tiny Emotion-6}&\makebox[4.8em]{\tiny WEBEmo} &\makebox[2.5em]{\tiny Self}  &\makebox[4.7em]{\tiny Mean Others}
		\\\hline\hline
		\tiny Deep Sentiment~\cite{you2015robust} & \scriptsize 78.74 & \scriptsize 68.38  & \scriptsize  49.76  &\scriptsize 47.79 &\scriptsize \textcolor{red}{78.74} &\scriptsize \textcolor{blue}{55.31} \\\hline
		\tiny Deep Emotion~\cite{you2016building} & \scriptsize  61.41 & \scriptsize 84.81  & \scriptsize 69.22 &\scriptsize 59.95 &\scriptsize \textcolor{red}{84.81} &\scriptsize \textcolor{blue}{63.52} \\\hline
		\tiny Emotion-6 (Sec.~\ref{sec:Bias}) & \scriptsize 54.33 &\scriptsize 64.28  &\scriptsize 77.72 &\scriptsize 64.30 &\scriptsize \textcolor{red}{77.72} &\scriptsize \textcolor{blue}{62.30}\\\hline
		\tiny WEBEmo (Ours) & \scriptsize  68.50 &\scriptsize  71.42 &\scriptsize  78.38&\scriptsize  81.41&\scriptsize \textbf{\textcolor{red}{81.41}} &\scriptsize \textbf{\textcolor{blue}{72.76}}\\\hline
	\end{tabular} 
\end{table*}

\noindent\textbf{Experiment 2: Correlation Analysis with Object/Scene Categories.}
Figure~\ref{fig:Obj_Scene1} shows the correlation between emotion and object/scene categories in our \textbf{WEBEmo} dataset. As can be seen from Figure~\ref{fig:Obj_Scene1}.a, less than 10\% of object categories are within the entropy range [0,0.6] for \textit{sadness} emotion leading to a much less biased dataset. This result is also consistent with the performance of the classifier trained for sadness vs non-sadness image classification in previous experiment (see Table~\ref{tab:neg}). We also observe that more number of scene categories have entropy in the higher range (see Figure~\ref{fig:Obj_Scene1}.b) showing that most of the scenes are well distributed across positive and negative emotion sets in our dataset.
Note that the negative bias still persists regardless of the large size of our dataset covering a wide variety of concepts (some object/scene categories still have zero entropy). 
We can further minimize the bias by adding weakly labeled images associated with zero entropy categories such that both positive and negative set can have a balanced distribution. This experiment demonstrates that our correlation analysis can help to detect as well as reduce biases in datasets.

\noindent\textbf{Experiment 3: Binary Cross-Dataset Generalization.}
Table~\ref{tab:cross1} summarizes the results. We have the following key observations from Table~\ref{tab:cross1}: (1) Model trained using our \textbf{WEBEmo} dataset shows the best generalization ability compared to the models trained using manually labeled emotion datasets. We believe this is because learning by utilizing web data helps in minimizing the dataset biases by covering a wide variety of emotion concepts. (2) More interestingly, on Emotion-6 dataset, the model trained using our stock images even outperforms the model trained with images from the same Emotion-6 dataset (77.72\% vs 78.38\%). This is quite remarkable as our model has only been trained using the web images without any strong supervision.

\textbf{Exploration Study.} To better understand effectiveness of curriculum guided learning strategy, we analyze cross-dataset generalization performance by comparing with following methods: (1) Direct Learning -- directly learning using the noisy web images of 25 fine-grained emotion categories, as in~\cite{you2016building,krause2016unreasonable,joulin2016learning}; (2) Self-Directed Learning -- start learning with a small clean set (500 images) and then progressively adapt the model by refining the noisy web data, as in~\cite{you2015robust,gan2016you}; (3) Joint Learning -- simultaneously learning with all the tasks in a multi-task setting. For details please refer to our supplementary material. We have the following key observations from Table~\ref{tab:expo}: (1) Performance of direct learning baseline is much worse compared to our curriculum guided learning. This is not surprising since emotions are highly complex and ambigious that directly learning models to categorize such finegrained details fails to learn discriminative features. (2) Self-directed learning shows better generalization compared to the direct learning but still suffers from the requirement of initial labeled data. (3) The joint learning baseline is more competitive since it learns a shared representation from multiple tasks. However, the curriculum guided learning still outperforms it in terms generalization across other datasets (70.38\% vs 72.76\%).  We believe this is because by ordering training from easy to difficult in a sequential manner, it is able to learn more discriminative feature for recognizing complex emotions.

\begin{table*} [t]
	\scriptsize
	\centering
	\caption{\scriptsize Exploration study on different webly supervised learning strategies.} 
	\begin{tabular}{|l||*{6}{c|}}\hline
		\label{tab:expo}
		\centering
		\tiny
		%\backslashbox{Train on:}{Test on:}
		Methods &\makebox[6em]{\tiny Deep Sentiment} &\makebox[5.4em]{\tiny Deep Emotion}  &\makebox[4.2em]{\tiny Emotion-6}&\makebox[4.8em]{\tiny WEBEmo} &\makebox[2.5em]{\tiny Self}  & \makebox[4.7em]{\tiny Mean Others}
		\\\hline\hline
		\tiny Direct Learning & 62.20&  67.48  &  74.73 & 76.65 & \textcolor{red}{76.65} & \textcolor{blue}{68.13} \\\hline
		\tiny Self-Directed Learning & 64.56 &\scriptsize  68.76  &\scriptsize 76.15 &\scriptsize  78.69&\scriptsize \textcolor{red}{78.69} &\scriptsize \textcolor{blue}{69.82}\\\hline
		\tiny Joint Learning & 66.71&\scriptsize  69.08  &\scriptsize  75.36 &\scriptsize  78.27 &\scriptsize \textcolor{red}{78.27} &\scriptsize \textcolor{blue}{70.38}\\\hline
		\tiny Curriculum Learning & 68.50 &\scriptsize  71.42 &\scriptsize  78.38&\scriptsize  81.41&\scriptsize \textbf{\textcolor{red}{81.41} }&\scriptsize \textbf{\textcolor{blue}{72.76}}\\\hline
	\end{tabular} 
\end{table*}

\textbf{Impact of Emotion Categories.} We compare our three stage curriculum learning strategy (2-6-25) with a two stage one involving only six emotion categories (2-6). We found that the later produces inferior results, with an accuracy of 78.21\% on the self test set and a mean accuracy of 70.05\% on other two datasets, compared to 81.41\% and 72.76\% respectively by the three stage curriculum learning. Similarly, there is a drop of 2.31\% in \enquote{self} test accuracy of the direct learning baseline while training with six emotion categories compared to the training with 25 emotion categories. In summary, we observe that the generalization ability of learned models increase with increased number of fine-grained emotion categories.

\textbf{State-of-the-Art Results.} Note that all the numbers presented in Table~\ref{tab:cross1} represent the binary accuracies we achieved without using any ground truth training data from the testing dataset. By fine-tuning, our model achieves a state-of-the-art accuracy of 61.13\% in classifying eight emotions on Deep Emotion dataset~\cite{you2016building} and an accuracy of 54.90\% on Emotion-6 dataset. Similarly, by utilizing training data from Deep Sentiment dataset, our model achieves an accuracy of 82.67\% which is about 8\% improvement over the prior work~\cite{you2015robust}.

\subsection{Analyzing Effectiveness of Our Learned Emotion Features}
\label{sec:features}

\begin{figure} [t]
	\centering
	\begin{tabular}{c}
		\includegraphics[scale=0.12]{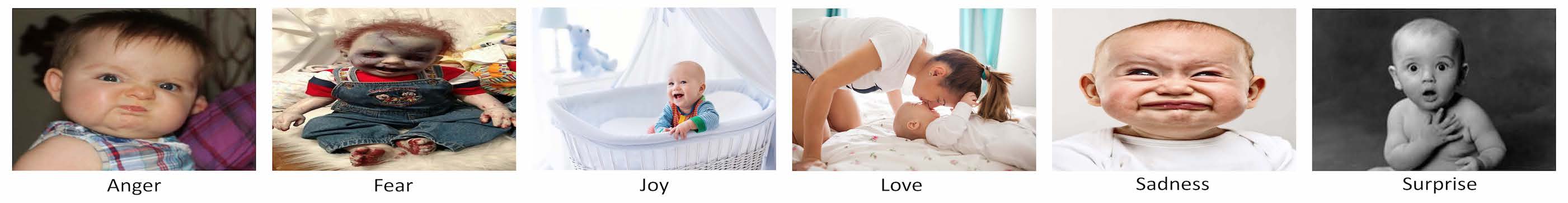}
	\end{tabular} 
	\caption
	{\scriptsize Sample images from our challenging \textbf{UnBiasedEmo} test set.See supplementary file for more example images on different object/scenes. Best viewed in color.}
	\label{fig:Baby} %\vspace{-1mm}
\end{figure}

\begin{SCtable}
	\label{tab:Fine} 
	\caption{\scriptsize Experimental results on our \textbf{UnBiasedEmo} test dataset. Features learned using curriculum learning outperforms all other basline features, including ImageNet.}
	\scriptsize
	\begin{tabulary}{1.1\linewidth}{|p{30mm}|P{23mm}|}
		\hline
		\textbf{Methods} & \textbf{Accuracy (\%)}   \\
		\hline
		ImageNet	&  64.20\\
		Direct Learning & 71.64\\
		Self-Directed Learning & 72.45\\
		Joint Learning	&  71.64\\
		Curriculum Learning	&  \textbf{74.27}\\
		\hline 		
	\end{tabulary} 
\end{SCtable}

\textbf{Experiment 1: Testing on Cross-Domain Unbiased Data.} In this experiment, we introduce a new unbiased emotion test set, \textbf{UnBiasedEmo} of about 3000 images dowloaded from Google to evaluate our learned models in recognizing very challenging emotions, e.g., different emotions with
same object/scene (see Figure~\ref{fig:Baby}). Since source of this test set is different from our \textbf{WEBEmo} dataset, it helps us alleviate the dataset bias issue in evaluation, so we can compare the generalization ability of various learning strategies in a less biased manner. Note that developing a large-scale unbiased dataset containing hundred thousands of images like this is a very difficult task as it requires extensive effort and also provides poor scalability. For an example, we could only able to get 3045 emotional images across six emotion categories (same as Emotion-6 dataset) from a collection of about 60,000 images.
More details on this unbiased dataset collection and annotations are included in the supplementary.

We use our learned models as feature extractors. We use 80\% of the images for training and keep rest 20\% for testing.
Table~\ref{tab:Fine} shows the classification accuracies achieved by the features learned using different methods. We have the following observations from Table~\ref{tab:Fine}: (1) Our curriculum learning strategy significantly outperforms all other baselines in recognizing fine-grained emotions from natural images. 
%(see Figure~\ref{fig:Qual} for some qualitative predictions by our model). 
(2) Among the alternatives, self-directed learning baseline is the most competitive. However, our approach still outperforms it due to the fact that we use the emotion hierarchy to learn discriminative features by focusing tasks in a sequential manner. (3) Performance of ImageNet features is much worse compared to the features learned using our curriculum guided webly supervised learning (64.20\% vs 74.27\%). This is expected as ImageNet features are tailored towards object/scene classification while emotions are more fine-grained and can be orthogonal to object/scene category, as shown in Figure~\ref{fig:Baby}.

We also inverstigate the quality of features learned using the current largest Deep Emotion dataset~\cite{you2016building} in recognizing image emotions on this unbiased test set and found that it produces inferior results, with an accuracy of 68.88\% compared to 74.27\% by our curriculum guided webly-supervised learning strategy on the \textbf{WEBEmo} dataset. We believe this is because of the effective utilization of large scale web data covering a wide variety of emotion concepts.

\begin{table}
	\parbox{.45\linewidth}{
		\centering
		\label{tab:}
		\scriptsize
		\begin{tabulary}{1.1\linewidth}{|p{30mm}|P{23mm}|}
			\hline
			\textbf{Methods} & \textbf{Accuracy (\%)}   \\
			\hline
			ImageNet	&  23.42\\
			Direct Learning & 25.43\\
			Self-Directed Learning & 24.92\\
			Joint Learning &  26.18\\
			Curriculum Learning	&  \textbf{27.96}\\
			\hline 		
		\end{tabulary} 
		\caption{\scriptsize Experimental results on Image Advertisement dataset. Our curriculum learning model performs the best.} }
	\hfill
	\parbox{.5\linewidth}{
		\centering
		\label{tab:}
		\scriptsize
		\begin{tabulary}{1.1\linewidth}{|p{32mm}|P{23mm}|}
			\hline
			\textbf{Methods}  & \textbf{Accuracy (\%)}   \\
			\hline
			ImageNet	& 43.27\\
			Direct Learning & 45.67\\
			Self-Directed Learning & 46.18\\
			Joint Learning	&   47.25\\
			Knowledge Transfer~\cite{xu2016heterogeneous}	& 45.10\\
			Curriculum Learning	&  \textbf{49.22} \\
			\hline 		
		\end{tabulary} 
		\caption{\scriptsize Experimental results on VideoStory-P14 dataset. 
			Features learned using our proposed curriculum learning outperforms the knowledge transfer approach by a margin of about 4\%.} 
	} 
\end{table}

\noindent\textbf{Experiment 2: Sentiment Analysis.}
We perform this experiment to verify the effectivenss of our features in recognizing sentiments from online advertisement images. 
We conduct experiments using Image Advertisement dataset ~\cite{hussain2017automatic} consisting of 30,340 online ad images labeled with 30 sentiment categories (e.g., active, alarmed, feminine, etc, -- see~\cite{hussain2017automatic} for more details).
We use the model weights as initialization and fine-tune the weights~\cite{hussain2017automatic}. We use 2403 images for testing and rest for training as in~\cite{hussain2017automatic}. We follow~\cite{hussain2017automatic} and chose the most frequent sentiment as the ground-truth label for each advertisement image.

Table 6 shows results of different methods on predicting image sentiments on the Advertisement dataset. From Table 6, the following observations can be made: (1) Once again, our curriculum guided learning significantly outperforms all other baselines in predicting sentiments from online ad images. (2) We achieve an improvement of about 6\% over the ImageNet baseline showing the advantage of our learned features in automatic ad understanding tasks.

\noindent\textbf{Experiment 3: Video Emotion Recognition.}
The goal of this experiment is to evaluate quality of our features in recognizing emotions from user videos.
We conduct experiments on VideoStory-P14 emotion dataset~\cite{xu2016heterogeneous} consisting of 626 user videos across Plutchik's 14 emotion classes.
We fine-tune the weights using video datasets and use 80\%/20\% of the videos in each category for training/testing. 
To produce predictions for an entire video, we average the frame-level predictions of 20 frames which are randomly selected from the video.

From Table 7, the following observations can be made: (1) We can see that all the models trained using \textbf{WEBEmo} dataset outperforms both ImageNet and transfer encoding features~\cite{xu2016heterogeneous} indicating the generalizability of our learned features in recognizing video emotions. (2) We further observe that curriculum guided learning provides about 2\% improvement over the joint learning baseline. 

\noindent\textbf{Experiment 4: Video Summarization.}
Our goal in this experiment is to see whether our learned features can benefit summarization algorithms in extracting high quality summaries from user videos.
We believe this is possible since an accurate summary should keep emotional content conveyed by the original video.

We perform experiments on the CoSum dataset~\cite{chu2015video} containing 51 videos covering 10 topics from the SumMe benchmark~\cite{gygli2014creating}. We follow~\cite{panda2017collaborative,chu2015video} and segment the videos into multiple non-uniform shots for processing.
We first extract pool5 features from the network trained with curriculum learning on our \textbf{WEBEmo} dataset and then use temporal mean pooling to compute a single shot-level feature vector, following~\cite{panda2017collaborative}. We follow the exact same parameter settings of~\cite{panda2017collaborative} and compare the summarization results by only replacing the visual features.

By using our learned emotion features, the top-5 mAP score of the recent summarization method~\cite{panda2017collaborative} improves by a margin of about 3\% over the C3D features~\cite{tran2015learning} (68.7\% vs 71.2\%). This improvement is attributed to the fact that good summary should be succinct but also provide good coverage of the original video's emotion content. This is an important finding in our work and we believe this can largely benefit researchers working in video summarization to consider the importance of emotion while generating good quality video summaries.

\noindent\textbf{Additional Experiments in Supplementary.} We analyze the effectivenss of our learned features in predicting communicative intents from persuasive images (e.g., politician photos)~\cite{joo2014visual} and see that our approach outperforms all other baselines by a signifcant margin ($\sim$8\% improvement over ImageNet features). We also provide sample prediction results in the supplementary material.

\section{Conclusion}
\label{sec:Conclusion}

In this paper, we have provided a thorough analysis of the existing emotion benchmarks and studied the problem of learning recognition models directly using web data without any human annotations. We introduced a new large-scale image emotion dataset containing about 268,000 high-quality images crawled from a stock website to train generalizable recognition models. We then proposed a simple actionable curriculum guided training strategy for learning discriminative emotion features that holds a lot of promise on a wide variety of visual emotion understanding tasks.
Finally, we demonstrated that our learned emotion features can improve state-of-the-art methods for video summarization. 

\textbf{Acknowledgements.} This work is partially supported by NSF grant 1724341 and gifts from Adobe. 
We thank Victor Hill of UCR CS for setting up the computing infrastructure used in this work.

\bibliographystyle{splncs04}
\bibliography{egbib}
\end{document}